\documentclass{article}
\usepackage{soul}
\usepackage{ijcai24}
\usepackage{xcolor}
\usepackage{times}
\usepackage{soul}
\usepackage{url}
\usepackage[hidelinks]{hyperref}
\usepackage[utf8]{inputenc}
\usepackage[small]{caption}
\usepackage{graphicx}
\usepackage{amsmath}
\usepackage{amsthm}
\usepackage{booktabs}
\usepackage{algorithm}
\usepackage{algorithmic}
\usepackage[switch]{lineno}

\usepackage[normalem]{ulem}
\title{Beyond the Limits: A Survey of Techniques to Extend the Context Length in Large Language Models}
\author{
Xindi Wang$^{1,2,3}$
\and
Mahsa Salmani$^1$\and
Parsa Omidi$^{1}$\and
Xiangyu Ren$^1$\and \\
Mehdi Rezagholizadeh$^1$\And
Armaghan Eshaghi$^1$\thanks{Corresponding author}
\\
\affiliations
$^1$Huawei Technologies Canada, Canada\\
$^2$University of Western Ontario, Canada \\
$^3$Vector Institute for Artificial Intelligence, Canada\\
\emails
xwang842@uwo.ca,
\{mahsa.salmani1, 
parsa.omidi,
xiangyu.ren1,
mehdi.rezagholizadeh,
armaghan.eshaghi\}@huawei.com
}
\begin{document}

\maketitle
%%%%%%%%%%%%%%%% Xindi %%%%%%%%%%%%%%%%%
\begin{abstract}
Recently, large language models (LLMs) have shown remarkable capabilities including understanding context, engaging in logical reasoning, and generating responses. However, this is achieved at the expense of stringent computational and memory requirements, hindering their ability to effectively support long input sequences. %However, LLMs are usually constrained by limitations in sequence length, and the expansion of sequence length represents a crucial requirement.
%in the era of large language models. 
This survey provides an inclusive review of the recent techniques and methods devised to extend the sequence length in LLMs, thereby enhancing their capacity for long-context understanding. In particular, we review and categorize a wide range of %approaches, including length extrapolation,  compression efficiency, attention-free approaches, attention approximation, and hardware-aware transformers. 
techniques including architectural modifications, such as modified positional encoding and altered attention mechanisms, which are designed to enhance the processing of longer sequences while avoiding a proportional increase in computational requirements. %Furthermore, we explore training, inference, and optimization strategies that have been instrumental in equipping LLMs with the ability to manage extended sequences in an efficient manner.
%Furthermore, we explore methods can be applied in 
The diverse methodologies investigated in this study can be leveraged across different phases of LLMs, i.e., training, fine-tuning and inference. This enables LLMs to efficiently process extended sequences.
The limitations of the current methodologies is discussed in the last section along with the suggestions for future research directions, underscoring the importance of sequence length in the continued advancement of LLMs.
\end{abstract}

%%%%%%%%%%%%%%%% Xindi %%%%%%%%%%%%%%%%%
\section{Introduction} 
In the rapidly evolving domain of natural language processing (NLP), large language models (LLMs), such as GPT-3, PaLM and LLaMA, emerged as pivotal tools that have proved proficiency %\MS{proficient} 
in understanding and generating human language including tasks such as language understanding, language generation, complex reasoning and other domains such as computer vision and autonomous driving \cite{NEURIPS2020_1457c0d6,touvron2023llama,chowdhery2023palm,wang2023investigating}.
%\cite{wang2023visionllm,wang2023investigating}. %,yang2023llm4drive}.
%These models demonstrate supreme performances across numerous tasks including language understanding, language generation, complex reasoning and other domains such as computer vision and autonomous driving \cite{wang2023visionllm,wang2023investigating,yang2023llm4drive}. %The remarkable advancements in performance achieved by LLMs can be attributed to their billions or even trillions of parameters, and they are trained on an extensive corpus of data aggregated from a variety of sources. 
% why process long sequences important?
%In numerous practical applications (e.g., multi-turn conversations), however, LLMs are expected to understand and/or generate very long context sequence so as to execute the task accurately in the inference phase. These context sequences are often significantly longer than that the LLMs have been pre-trained with. That highlights the fact that the LLMs must have the capability to deal with long sequences.
In many real-world scenarios, such as multi-turn conversations and document summarization, LLMs are required to comprehend and produce long sequences in order to perform the task accurately during the inference phase. These context sequences are often substantially longer than those the LLMs were trained with, emphasising the fact that LLMs must have the capability to deal with lengthy sequences.
%In numerous practical applications, for instance, multi-turn conversations, LLMs frequently encounter the necessity to understand or produce context sequences that are significantly longer than what they have been pre-trained with, which requires the model to have the capability to deal with long sequences.
%Nonetheless, LLMs encounter various restrictions and challenges that impede them to support long sequences efficiently.
%Although LLMs are spearheading the forthcoming wave of AI revolution, %a significant challenge that persists in impeding 
%their capabilities is the restricted capacity to process long sequences efficiently.} 
%These constraints not only impact the coherence and contextual relevance of the generated sequences but also restrict the effectiveness of LLMs in complex linguistic tasks that require a comprehensive grasp of context. The capability of dealing with long sequences is pivotal in tasks such as document summarization, extended conversation handling, and complex question-answering, where context plays a crucial role in achieving accuracy and relevance.

% chanllenges for long sequences
Processing long sequences by LLMs is a non-trivial task, which involves computational, structural, and practical challenges. Notably,
%\sout{the surge in the need for computational resources, a consequence of} 
increased sequence lengths can exponentially escalate processing requirements, particularly in transformer-based models with self-attention mechanisms. %\sout{This not only increases} \MS{Such significant increase in the processing requirements comprises both computational complexity and memory demands, which} 
This not only increases the computational cost but also, the memory demands often surpass the capacity of advanced GPUs and thus, impeding efficient training~\cite{dao2022flashattention}. Hence, the efficiency of attention mechanisms, pivotal in addressing longer sequences, remains a key area of research, aiming to balance computational efficiency with model performance~\cite{gu2023mamba}. Moreover, %\sout{the task of}
maintaining contextual understanding and coherence over extended input spans further complicates the scenario, as it requires advanced methods to capture and utilize long-range dependencies. Finally, the evaluation and benchmarking of LLMs on long-sequence tasks also pose a significant challenge, demanding novel metrics and datasets for effective assessment~\cite{kwan2023m4le}. Altogether, the aforementioned challenges highlight the intricacy and importance of advancing LLMs to proficiently support and utilize long sequences for various tasks. 

\begin{figure*}[ht]
\includegraphics[width=\textwidth]{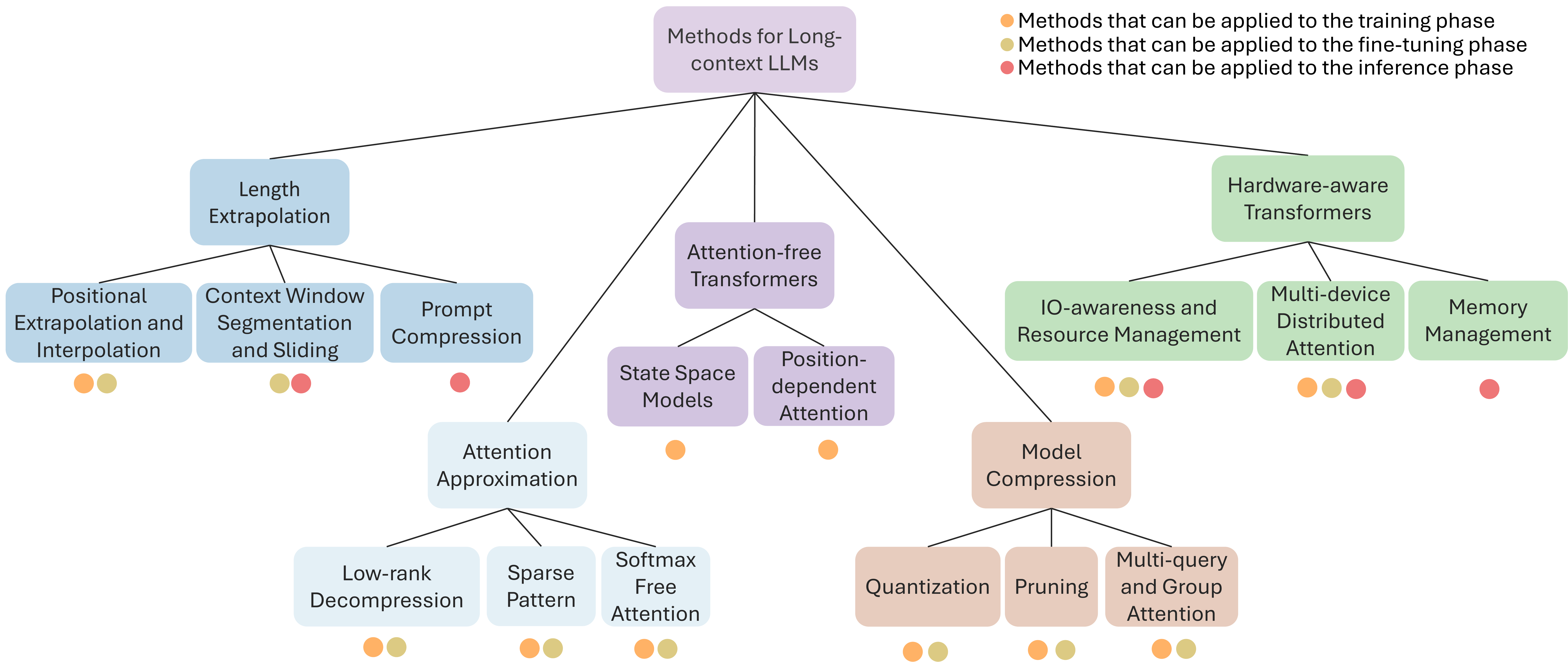}
\caption{Taxonomy of Long-context %\sout{Large Language Model} \mehdi{LLM} 
LLM literature, which includes five distinct sections: length extrapolation, attention approximation, attention-free transformers, model compression, and hardware-aware transformers. We also establish connections between the methodologies and their related applicability scenarios. Some entail training a new model from scratch, others involve fine-tuning %\sout{pre-existing} \mehdi{pre-trained} 
pre-trained models, and some implement over inference without any updates of hyper-parameters. %\mehdi{The colors are not distinguishable in the printed version and the fonts are too small (not easy to read). }\Xindi{I will update the figure once we have settled down with the section names and order.}
}
\label{fig:1}
\end{figure*}
% why this survey 
In this survey, we provide a %sout{comprehensive} \MS{
concise review of various approaches that have been developed to enable LLMs to handle long sequences. %\mehdi{you can just say "long sequences"}. 
The overarching goal of the survey is to provide a detailed insight into those methods, as well as to highlight possible directions for future research. %, including length extrapolation,  compression efficiency, attention-free approaches, attention approximation, and hardware-aware transformers. 
The techniques include architectural modifications, such as positional encoding modification, modified attention mechanisms and model compression techniques, which aim to optimize the processing of longer sequences without exponentially increasing computational and memory demands.  Additionally, we explore the methods that can be adopted in different phases (training, fine-tuning, and inference), and have been pivotal in enabling LLMs to handle longer sequences, efficiently. The taxonomy of our literature review is shown in Figure \ref{fig:1}. While there are existing surveys addressing LLMs with a more general scope \cite{zhao2023survey,naveed2023comprehensive,wan2023efficient}, this survey is particularly focused on evaluating the articles dealing with long sequences in LLMs. %\sout{Additionally,} \MS{
Moreover, there are other reviews on efficient Transformers and their training methodologies \cite{ijcai2023p0764,huang2023advancing}, but this survey specifically focuses on %\sout{efficiency techniques tailored for} 
models and strategies that aim at enhancing the management of longer input sequences.

\section{Length Extrapolation} %\Mahsa{Length Extrapolating Methods?}}
%\Mahsa{
In this section, %we delve into the initial category of approaches designed to enhance the capability of LLMs to handle long sequences. In particular, 
we focus on methods whose primary objective is to enable LLMs to effectively support longer sequences. Among these methods, positional extrapolation and interpolation emerge as pivotal methods for extending the model's capacity to handle sequences longer than those on which the LLMs have been originally trained.
%In the advancing field of LLMs, the ability to effectively process and interpret long sequences is crucial for enhancing their performance and applicability. 
%\sout{This section delves into techniques that aim to directly enhance the efficacy of LLMs in processing lengthy. Among these, positional extrapolation and interpolation stand out as key methods for extending the model's ability to manage sequences beyond learned lengths.} 
Furthermore, we explore context window segmentation and sliding, a crucial technique that manipulates input sequences into smaller segments or moves the context window to enable processing of the longer sequence. Lastly, we review the strategy of prompt compression, an innovative approach to condense input prompts efficiently while retaining the essential information.

\paragraph{Positional Extrapolation and Interpolation.} Position extrapolation and interpolation refer to the techniques that adjust the positional embedding (PE) associated with input tokens, which modify how these tokens are positioned and interpreted within the model's architecture. PEs play a pivotal role in the architecture of transformer models since they impart a crucial sense to the input tokens, enabling the model to discern the specific position of each token within the sequence. This ensures that the model can effectively capture and utilize the sequential information inherent in the input data. The vanilla transformer \cite{NIPS2017_3f5ee243} presents a novel Sinusoidal PE (SinPE) that uses sinusoidal functions to represent the absolute positions of the tokens. %While SinPE is a widely used method, recent research has explored alternative approaches to handle positional information in transformer models. 
SinPE has become a widely used method, yet it has prompted further research into alternative approaches for handling positional information in transformer models. One alternative approach is trainable PEs, as explored by Chen {\em et al.}~\shortcite{chen-etal-2021-simple}, which learn an embedding mapping specific to the task. Another approach focuses on relative PEs, introduced by Shaw {\em et al.}~\shortcite{shaw-etal-2018-self}, which encodes the relative positions of tokens rather than their absolute positions, allowing for more flexible handling of varying sequence lengths. Additionally, the concept of Rotary PEs (RoPE)~\cite{su2024roformer}, involves rotating the query and key representations at an angle corresponding to the absolute positions of the tokens within the input sequence. This method provides a unique way of integrating positional information that can enhance the model's ability to capture complex dependencies. To further improve efficiency and support longer sequences, recent studies have investigated methods for positional extrapolation and interpolation. 

%For instance, trainable PEs \cite{chen-etal-2021-simple} involve learning an embedding mapping, while relative PEs \cite{shaw-etal-2018-self} are based on relative positions. Rotary PEs (RoPE) \cite{su2023roformer} are the PEs that rotate the query and key representations %\Mahsa{matrices?} 
%at an angle which corresponds to the absolute position of the tokens within the input sequence. PEs can be further modified to provide higher efficiency in supporting longer sequences. In this work, we review the methods related to positional extrapolation and interpolation that can improve the PEs. These techniques address the inherent limitations of fixed sequence lengths, pushing the boundaries of what LLMs can comprehend and produce.

%\paragraph{Positional Extrapolation} 
Positional extrapolation refers to the model's ability to handle input sequences %\Mahsa{
that exceed the length of %\sout{longer than} 
those it was trained on, enabling %\Mahsa{
the preservation of %\sout{it to maintain} 
context and coherence over extended sequences. This capability is important for models tasked with understanding and generating lengthy documents or conversations. For example, Attention with Linear Biases (ALiBi)~\cite{press2022train} introduces a heuristic of negative causal attention bias, which dispenses with PEs for tokens in the transformer model. ALiBi encodes position information by biasing the query-key attention scores proportionally to the distance between each pair of tokens. %\Mahsa{
As compared to other PE schemes, ALiBi %\sout{It} 
demonstrates superior extrapolation capabilities to unseen sequence lengths. %\sout{when compared to other PE schemes.} %Remarkably, ALiBi facilitates training on shorter sequences without experiencing performance deterioration, even when the inference sequence length significantly exceeds the training sequence length. This resilience extends from sequences 2 times longer to 8 times longer, and potentially beyond.
Different from ALiBi, xPOS \cite{sun-etal-2023-length} extends causal RoPE, which incorporates a unique exponential decay factor at each dimension of the rotation angle vector, thereby improving length extrapolation. Another approach CLEX~\cite{chen2024clex} uses ordinary differential equations to generalize PE scaling. By modeling continuous dynamics with length scaling factors, CLEX effectively overcomes the constraints of traditional positional extrapolation techniques.
%Despite using manually-designed scaling strategies, CLEX~\cite{chen2023clex} uses ordinary differential equations to generalize PE scaling. This method models the continuous dynamics using length scaling factors, which effectively overcome the constraints of traditional positional extrapolation scaling techniques. 

%\paragraph{Positional Interpolation} 
On the other hand, positional interpolation deals with the model's proficiency in inserting or integrating new information within existing sequences. %, a crucial skill for tasks such as text editing and information updating. 
For example,  positional interpolation proposed by Chen {\em et al.}~\shortcite{chen2023extending} applies linear scaling on the position indices, effectively aligning the maximum position index to correspond with the context window limit previously established during the pre-training phase. Experimental observations indicate that this strategy exhibits greater stability and necessitates fewer fine-tuning steps compared to direct extrapolation methods.
Additionally, YaRN \cite{peng2023yarn} extends RoPE by adopting an uneven interpolation of frequencies, specifically preserving the high-frequency components. %, which avoids losing important positional details.
This approach avoids losing important positional details that enhances the ability of the model to maintain critical positional information.

%Length extrapolation refers to the techniques that are used to extend the generative capabilities of the LLMs, %The Length Extrapolation section explores methods to manipulate pre-trained models with limited sequence lengths, 
%which enables the LLMs to process longer sequences. Techniques such as  %\sout{, as well as} 

\paragraph{Context Window Segmentation and Sliding.} 
%Transformer-based 
LLMs based on transformers are inherently constrained by limited context windows, rendering them incapable of directly integrating or utilizing the entirety of information in long sequences.
%\sout{that surpass these context boundaries.} 
To mitigate this limitation, various methodologies have been developed to divide the input into segments and apply a sliding window approach to manage the context. 
One such approach is structured prompting~\cite{hao2022structured}, %\Mahsa{initially focused on scaling in-context learning to thousands of examples,} 
which groups demonstration examples and encodes them individually with well-designed position encoding. These encoded examples are then collectively attended to by the test example through a re-scaled attention mechanism, ensuring that each segment receives adequate focus and relevance.
%Subsequently, these examples undergo collective attention by the test example through a re-scaled attention mechanism.
Building on the idea of segmenting input, Ratner {\em et al.}~\shortcite{ratner-etal-2023-parallel} introduces a parallel context window (PCW), %that segments the long context into chunks, 
which segments the long-context into chunks and restricts the attention mechanism to operate exclusively within each window. By redeploying positional encoding across these windows, this method ensures efficient processing of long sequences without overwhelming the attention mechanism.
%restricting the attention mechanism to operate exclusively within each window. Subsequently, positional encoding is redeployed across these windows.
Another innovative approach is StreamingLLM \cite{xiao2023efficient}, which addresses the ``attention sink'' phenomenon. This phenomenon occurs when a significant portion of the attention score is allocated to the initial tokens, regardless of their relevance. StreamingLLM merges window context with the first token,
%observes the \textit{attention sink} phenomenon that a large amount of attention score is allocated to the initial tokens, irrespective of their relevance to the language modelling task. This observation suggests an efficient framework by merging window context and the first token, 
which enables the LLMs trained with a finite-length attention window to be effectively generalized to infinite sequence lengths without requiring additional fine-tuning. 

%%%%%%%%%%%%%%%% Xindi %%%%%%%%%%%%%%%%%
\paragraph{Prompt Compression.}
Prompt compression refers to methods that shorten original prompts while keeping the important information. %The process of prompt compression expedites the processing of LLM inputs by either condensing extensive prompt inputs or acquiring concise representations of prompts through learning. 
This process involves either condensing extensive prompt inputs or learning concise representations of prompts.
LLMLingua \cite{jiang-etal-2023-llmlingua} employs streamlined and proficient language models, such as GPT-2 small or LLaMA-7B, to identify and eliminate extraneous tokens within prompts. This method facilitates the efficient execution of inferences with expansive language models, achieving a compression ratio of up to 20 times while maintaining performance with minimal decline. Building on this approach, LongLLMLingua \cite{jiang2023longllmlingua} addresses the inherent ``lost in the middle'' issue observed in LLMs, enhancing the processing of long-context information. This method not only reduces costs but also improves efficiency through prompt compression, resulting in a significant improvement of up to 21.4\% in retrieval-augmented generation performance while using only a quarter of the tokens.
Further advancing the field, Li {\em et al.}~\shortcite{li-etal-2023-compressing} introduce a novel method called ``Selective Context''. %that aims to augment the inference efficiency of LLMs by systematically identifying and pruning redundancy within the input context. The primary objective is to streamline the input, rendering it more compact and thereby optimizing the overall efficiency of language model inferences.
This approach systematically identifies and prunes redundancy within the input context to streamline the input, making it more compact and optimizing the overall efficiency of language model inferences.
%%%%%%%%%%%% Parsa %%%%%%%%%%%%%%%%
MemGPT \cite{packer2023memgpt} is then proposed to 
%is another prompt processing-based model that aims to 
overcome the limitations of fixed-length context windows in traditional LLMs. The primary goal is to simulate an infinite context while still effectively utilizing fixed-context models. MemGPT achieves this by autonomously managing its own memory through ``function calls" allowing for dynamic context modifications during a task. It establishes a memory hierarchy, akin to traditional operation systems, and treats context windows as constrained memory resources. By enabling the LLM to control its context, MemGPT provides an illusion of longer context length.
%%%%%%%%%%%%%%%%%%%%%%%%%%%%%%%%%%%%%%%%%%%%%%%%%%%%%%%%%%%%%%%%%%%%%%
%%%%%%%%%%%%%%%%%%%%%%%%%%%%%%%%%%%%%%%%%%%%%%%%%%%%%%%%%%%%%%%%%%%%%%%%%%%%%%%%%%%%%%%%%%%%%%%%%%%%%%%%%%%%%%%%%%%%%%%%%%%%%%%%%%%%%%%%%%%%
\section{Attention Approximation}
% transition paragraph is needed here!
The foundation of attention approximation lies in the ambition to reduce the computation and memory complexities of vanilla self-attention~\cite{NIPS2017_3f5ee243}, which increases quadratically with respect to the sequence length $n$, i.e., $O(n^2)$. This can be achieved by approximating the full-rank attention map with a low-rank counterpart, exploiting the sparse patterns in the attention layers, or %deconstructing the softmax-related complexity of vanilla attention to obtain a simplified yet efficient approximation. 
simplifying the softmax-related complexity of vanilla attention. These techniques aim to provide efficient approximations that maintain the effectiveness of the attention mechanism while managing long sequences more efficiently.

%%%%%%%%%%%%%%%% Moshi %%%%%%%%%%%%%%%%%
\paragraph{Low-rank Decomposition.}
The transformer architecture utilizes a self-attention mechanism that involves three matrices, namely, Query ($\mathbf{Q}$), Key ($\mathbf{K}$), and Value ($\mathbf{V}$). The attention mechanism works by computing the similarity between the $\mathbf{Q}$ and $\mathbf{K}$ %matrices, %($\mathbf{Q}\mathbf{K}^T$), 
and the result is used to weight the $\mathbf{V}$, emphasizing the the most relevant information. %i.e., \big($\operatorname{softmax}(\mathbf{Q}\mathbf{K}^T)\mathbf{V}$\big). 
The low-rank decomposition method can make the attention computation more efficient by reducing the number of parameters in the matrices. 
% This method decomposes the attention mechanism of $\mathbf{Q, K}$ and $\mathbf{V}$ into smaller matrices. % The classic approach for Low-rank decomposition is to decompose the matrix using singular value decomposition(SVD). 
%Low-rank decomposition has been widely studied in the literature. 
One such approach is Linear Encoder-Decoder (LED)~\cite{winata2020lightweight}, %algorithm 
which is proposed to decompose each of the three matrices into smaller matrices by adding an encoder and decoder before and after the self-dot-product to reduce the matrix size for approximation of linear parameter efficiency. Different from LED, Linformer~\cite{wang2020linformer} introduces another linear projection mechanism that adds two smaller matrices
%$\mathbf{E}$ and $\mathbf{F}$ 
before $\mathbf{K}$ and $\mathbf{V}$ to project them to a smaller size while leaving $\mathbf{Q}$ unchanged. Both methods optimize matrix computation through linear approximation. Autoformer~\cite{NEURIPS2021_bcc0d400} further improves the ability of capturing long-term dependency by introducing an auto-correlation mechanism that leverages the Fast Fourier Transform (FFT) for time series decomposition. The decomposed matrix is then utilized for time series analysis, which enables the model to better capture and improve forecasting accuracy for long-term contexts. Deep neural networks (DNNs) have also been utilized for tensor decomposition in transformers. In particular, unlike traditional methods such as singular value decomposition, Deeptensor~\cite{saragadam2022deeptensor} uses a DNN to learn an optimal regularizer for tensor decomposition when the distributions of the tensor is non-Gaussian. 

\paragraph{Sparse Pattern.}
%As mentioned above, the computation complexity and memory usage of Transformers increases quadratically with respect to the sequence length $n$, i.e., $O(n^2)$. Such stringent memory and computational requirements preclude Transformers from supporting long sequences. One of the main approaches 
%Another approach to address the computation and memory challenges in Transformers is to exploit sparse patterns. These patterns sparsify the connections in the attention layers, which result in a sparse attention matrix \cite{child2019generating}, \cite{yun2020n}. The sparse attention matrix indicates only a sparse set of tokens that each unique token should attend to. 
An alternative strategy to address the computation and memory challenges of the self-attention module in transformers involves leveraging sparse patterns to handle long contexts effectively. %These patterns induce sparsity in the connections within attention layers, leading to a sparse attention matrix~\cite{yun2020n}, %\cite{child2019generating,yun2020n}, 
%which illustrates a limited set of tokens that each unique token should attend to. 
These patterns use a sparse attention matrix, where each token attends to a limited set of other tokens.
Various methods have been proposed to introduce the sparsity, which, while not specifically designed for long contexts, can effectively help manage long sequences. %including the notion of 
%One approach is local attention, where only neighboring tokens attend to each other, 
%\sout{that proximate tokens should be attended} 
%while distant tokens may lose their connections, the introduction of randomness in connections, the assertion that a global token must attend to every token for comprehensive awareness, and combinations of these strategies.
%which may limit connections with distant tokens. Another method introduces randomness in connections to cover long-range dependencies, while a global token attends to every token to maintain comprehensive awareness across the entire sequence. Combinations of these strategies are employed to balance local and global context, ensuring efficient processing of long sequences.
%\mehdi{better to add citations for each category. }
%The attention sparse patterns can either be fixed, which only depends on the location of the tokens, or adaptive, where the pattern considered the embedding values as well to construct the attention matrix. 

%\Mahsa{
Among the most straightforward yet practical instances of sparse patterns, %the block-wise paradigm, e.g., 
Block-wise Self Attention~\cite{qiu-etal-2020-blockwise}, stands out as an illustrative demonstrations. This method reduces the computation and memory cost by chunking the input sequence into fixed blocks. An alternative strategy involves having an individual token attend to tokens at regular, fixed internals. For instance, Longformer~\cite{beltagy2020longformer} is a sparsifying mechanism that utilizes dilated windows of tokens to construct the attention matrix. LogSparse~\cite{NEURIPS2019_6775a063} is another method that sparsifies the attention matrix by restricting consideration to a limited window of tokens, %while in this approach, 
where the window is defined by exponential steps from the token itself. This approach ensures a targeted focus range for each individual token.
%by considering only a limited window of $n$ tokens, obtained by considering exponential steps from the token itself, for each specific token to attend to. \mehdi{The sentence is not clear. Better to be re-written.} 
%The authors show that by 
By employing LogSparse, it %can be 
is guaranteed that any pair of tokens can exchange attention information with each other, while the memory usage of the transformer can be reduced to $O(n(Log~n)^2)$. LongNet \cite{ding2023longnet} introduces dilated attention, in which attention allocation decreases exponentially as the distance between tokens increases. This approach exploits mixed dilated rates to accommodate both local and global dependencies between different tokens. It has been shown that by utilizing LongNet, a linear computation complexity, $O(n)$, and a logarithm dependency between tokens can be achieved.  

Some other sparse transformers consider adaptive sparse patterns which are not dependent on the location of the tokens, but rather they rely on other dynamic factors such as embedding values or task-specific parameters. For instance, Routing Transformer \cite{roy-etal-2021-efficient} exploits dynamic key-value pairs to infer sparsity patterns and hence, it removes the computation and memory requirements of attending to content unrelated to the query of interest. In particular, Routing Transformer utilizes $k$-means clustering to define the $k$ most relevant columns in $\mathbf{Q}$ and $\mathbf{K}$, and assigns each query to the keys within the same cluster. Routing Transformer results in computation complexity of the order $O(n^{1.5})$. %\Mahsa{
Reformer~\cite{Kitaev2020Reformer} is another sparse approach which clusters the tokens prior to implementing attention, and it does so according to a hash-based similarity.

%\mehdi{There are other very important papers to cover such as Blockwise Self Attention~\cite{qiu2020blockwise}, Reformer~\cite{kitaev2020reformer}, Longformer~\cite{beltagy2020longformer}, BigBird~\cite{zaheer2021big}, ...  }

%%%%%%%%%%%%%%%% Parsa %%%%%%%%%%%%%%%%%
\paragraph{Softmax-free Attention.}
The efficacy of vanilla attention~\cite{NIPS2017_3f5ee243} is often attributed to the softmax operation, which is %a vital element in 
important for capturing long dependencies. However, this operation %introduces 
causes quadratic complexity in both time and space, impeding the %seamless scaling up 
scalability of transformers %and/or having a long sequence of tokens as input. 
for long sequences. %The replacement of 
Replacing the softmax operation %has the potential to 
can reduce computational complexity, %paving the way for improved efficiency in
enhancing the efficiency of processing %extensive token 
long sequences. This category of approaches is called softmax-free attention.

CosFormer~\cite{zhen2022cosformer} emulates softmax behaviors
%, including non-negativity and nonlinearity, 
through a linear operator% employing a cosine-based distance re-weighting mechanism. 
that re-weights the cosine-based distance. 
SOFT~\cite{NEURIPS2021_b1d10e7b} employs a Gaussian kernel function to replace the softmax, %operation, 
while SIMA~\cite{10483688} opts for normalizing query and key matrices using a simple L1-norm. Another set of approaches replaces softmax with the ReLU function for normalization, demonstrating that this substitution maintains performance, while preserving linear scalability~\cite{shen2023study}. % ~\cite{hron2020infinite,shen2023study,bai2023transformers}.
%\Mahsa{
An alternative class of architectures centers around generalized kernelziable attention, wherein the conventional attention mechanism is formulated as a specific kernel function. For instance, Performer~\cite{choromanski2021rethinking} is an approach leveraging positive orthogonal random features to effectively model the attention mechanism into simplified softmax-free architecture with linear space and time complexity.

%\mehdi{There are some missing papers here such as LLN~\cite{nahshan2023linear}, TransNormerLLM~\cite{qin2024transnormerllm}, TransNormer~\cite{qin2022devil}, Performer~\cite{choromanski2022rethinking}, RFA~\cite{peng2021random}}

%\mehdi{I am not sure if RetNet should go here. It is more like a new architecture to me than a softmax-free attention.  }
%\Mahsa{
Another recently-developed transformer architecture that can be studied under this category (to varying degrees) is
%Another standout in this category is 
RetNet~\cite{sun2023retentive}, which replaces the softmax operation with a D-matrix followed by group normalization (GroupNorm). The D-matrix introduces exponential decay weighting of previous tokens, diminishing the impact of distant tokens. The incorporation of GroupNorm adds non-linearity, a characteristic once inherent in softmax. A distinguished feature of RetNet is that it can be implemented in both parallel and sequential manners. Accordingly, it can exploit the accelerated token generation during inference, similar to Recurrent Neural Networks (RNN), and exploit the efficiency of parallelization during training.

%%%%%%%%%%%%%%%% Parsa %%%%%%%%%%%%%%%%%
\section{Attention-free Transformers}
Attention-free transformers refer to the computational approaches that provide dependency information between tokens without relying on the conventional attention mechanism. 
These mechanisms offer a different perspective on dependency calculation, while maintaining sub-quadratic memory complexity. In this study, we consider two distinct sub-categories of this domain, namely, State Space Model (SSM) and positional-dependency attention—that enhance the handling of long contexts in LLMs.
\paragraph{State Space Model.}
SSM is a statistical sequence-to-sequence (seq2seq) model that employs linear projections of hidden states to compute the output sequence based on an input sequence. 
%~\cite{kalman1960new}. %~\cite{kalman1960new,gu2021combining}
%The use of 
SSM introduces an RNN-like seq2seq model without non-linearity, which %the absence of non-linearity 
empowers parallel training%, while the RNN structure exploits efficient token generation during inference. 
and optimize the inference efficiency. 
%Unlike traditional RNN, SSM employs a hidden state size on the order of hundreds of sequence lengths. 
The seq2seq operation based on the states can be analytically unrolled, resembling a convolutional operation with a parametrized kernel. Theoretically, similar to RNN, this convolution operation can extend to infinite length, enabling %output computation without the necessity to compute individual states. 
the computation of outputs without %the need to 
calculating individual states. During the training phase with the entire input sequence, this convolution process can be exceptionally rapid and parallel, setting it apart from the %characteristics of 
traditional RNN. However, the computational expense of the convolution kernel limits SSM's application in deep learning until the advent of Structured State Space (S4)~\cite{gu2022efficiently}.
S4 %is a combination of 
integrates SSM, HIPPO~\cite{NEURIPS2020_102f0bb6}, and structured  matrices to solve the complexity of the convolution kernel. HIPPOO~\cite{NEURIPS2020_102f0bb6} is a particular representation of the original SSM, which takes input states and maps them to higher-dimensional states that can be seen as an online compression of history. However, with a finite size of states, this method cannot remember the entire input. This necessitates the introduction of exponential decay, particularly beneficial for recent past accuracy. Some approaches %Works such as 
%~\cite{li2022approximation,orvieto2023resurrecting,ma2022mega} 
employ decay techniques, showcasing performance improvements~\cite{orvieto2023resurrecting,ma2023mega}.
Hungry Hungry Hippos (H3)~\cite{fu2023hungry} incorporates two SSMs, enabling local token attention and global token recall through a multiplicative gate mechanism, akin to LSTM gating. %One SSM's role involves providing a local convolution, allowing the model to revisit earlier tokens in the sequence. 
Hyena~\cite{poli2023hyena}, similar to H3, replaces the attention layer by interleaving implicitly parametrized long convolutions
and data-controlled gating, 
%an MLP-parameterized global convolution, 
effectively narrowing the quality gap with the vanilla attention mechanism at scale and achieving comparable perplexity with a reduced computational cost. Mamba~\cite{gu2023mamba} enhances SSMs by incorporating H3 with multi-layer perceptrons (MLP), refining reasoning capabilities through a strategic reorganization of the gating mechanism. %The time-varying nature of this model necessitates an innovative hardware-aware algorithm, leveraging the memory hierarchy for the efficient and rapid implementation of dynamic convolutions.

\paragraph{Position-dependent Attention.}
Within this distinct category, a unique form of dependency calculation emerges, where dependencies rely on the position of tokens rather than interactions between them. The Attention-free Transformer (AFT) ~\cite{zhai2021attention}, inspired by attention-based transformers, exclusively employs $\mathbf{K}$ and $\mathbf{V}$ while eliminating $\mathbf{Q}$ and its dot product with $\mathbf{K}$. %However, a novel learnable matrix, $\mathbf{W}$, is introduced, which, unlike the adaptive weighting matrix in vanilla attention, remains consistent across all input sequences, serving as a fixed attention map (static routing). 
Instead, AFT introduces a novel learnable matrix $\mathbf{W}$, which acts as a fixed attention map (static routing) consistent across all input sequences. 
Unlike adaptive weighting in vanilla attention, $\mathbf{W}$ considers only pairwise token positions, disregarding semantic dependencies. To enhance customization based on current input data, %key vector 
$\mathbf{K}$ accompanies $\mathbf{W}$. 

Building upon the principles of AFT, Receptance Weighted Key Value (RWKV) architecture~\cite{peng-etal-2023-rwkv} adopts a similar approach with modifications to interaction weights for simplicity, 
%modifies the interaction weights, 
and redefines $\mathbf{W}$ as a linear time decay of a vector with a much smaller size. RWKV provides the flexibility to formulate a seq2seq model as either a transformer or an RNN, similar to what we observe in RetNet.
%within the softmax-free subsection. 
This proves advantageous for parallelizing computations during training using the transformer form of RWKV while maintaining consistent computational and memory complexity during inference through the RNN form, without limitations on sequence length.
Although both AFT and RWKV imply trade-offs between performance and complexity, RWKV emerges as a practical alternative for dot-product transformers with the ability to scale up to very large models.

%%%%%%%%%%%%%%%%%%%%%%%%%%%%%%%%%%%%%%
\section{Model Compression}
An alternative approach that can enable LLMs %architectures 
to support longer sequences is model compression.
%In general, compression implies the methods that minimize either the size of LLM architecture or the memory and computation required by LLMs, while trying to maintain the performance and accuracy as much as possible. 
Various model compression approaches have distinct focal points. Some concentrate on minimizing the size of the LLM architecture by eliminating redundant weights, thereby reducing computational and memory requirements. Some others prioritize decreasing computation precision to alleviate computational complexity. Furthermore, certain approaches emphasize enhancing memory efficiency and optimizing data storage methods. 
In this section, we explore methods that exert a more significant impact on accommodating longer input sequences.
%, especially focusing on quantization, pruning,and multi-query approach.
%, multi-query approach, and prompt compression. 
%%%%%%%%%%%%%%%% Mahsa %%%%%%%%%%%%%%%%%
%%\subsection{Model Compression}
%There are various methods for model compression, namely, quantization, parameter pruning, low-rank approximation, and knowledge distillation. In this paper, we study the quantization, pruning, and knowledge distillation, the ones that play important roles in supporting longer sequences. 
\paragraph{Quantization.}
Quantization has been considered as a promising approach for improving the computational time and energy efficiency of generic neural networks.
%as it can alleviate the memory and computational overhead of the networks. 
Moreover, neural networks are robust enough to be quantized to lower bit-widths with a relatively small impact on the accuracy of the network \cite{gholami2022survey}. That provides an insight into utilizing quantization to reduce the complexity of LLMs, and accordingly, enabling them to support longer input sequence \cite{zhu2023survey}.
Depending on the stage at which quantization is implemented, quantization techniques for LLMs can be classified as Quantization-Aware Training (QAT) and Post-Training Quantification (PTQ).

In QAT approach, the quantization is integrated into the training phase such that %the quantization effects can be accommodated in the training procedure. Accordingly, 
the network can be adapted to quantization effects.
%and, hence, be more robust to the reduced precision. 
This adaptation helps mitigate the potential loss of accuracy that might occur as a result of quantization
%when moving from higher precision to lower precision 
during the inference phase. However, applying QAT to LLMs can be challenging due to the computational cost and the latency, as QAT requires training over the whole training dataset to avoid significant accuracy degradation. LLM-QAT \cite{liu2023llm} addresses this issue by proposing data-free knowledge-distillation, in which the data generated by the LLM itself is used for knowledge distillation. As the proposed approach can retain the distribution of the non-quantized (original) output, it can be applied to any generative model, independent of the original training dataset.

On the other hand, PTQ involves reducing the precision of the weights and activations of a neural network after the completion of the training phase. The primary goal of PTQ is to reduce the memory and computational requirements of the model, making it more suitable for deployment on resource-limited devices. PTQ is simple and efficient, however, it can impose performance degradation due to the low precision. With the existing trade-off between the model size, computation speed and accuracy, this method can be used to improve the efficiency of LLMs without extensive training efforts. 

The PTQ approaches can be categorized into weight-only quantization, which only focuses on quantizing the weights 
%and keeps the activations in full-precision representation, 
and weight-activation quantization, which quantizes both weights and activations. LLM.int8() \cite{NEURIPS2022_c3ba4962} is the first multi-billion-scale INT8 quantization procedure that reduces memory usage by half during inference, while it maintains the performance the same as that in the full-precision model. OPTQ \cite{frantar2023optq} proposes a layer-wise quantization technique, which can further reduce the precision to 3 or 4 bits per weight element, with negligible accuracy degradation.
%as compared to the non-quantized version. 
%The observation in \cite{frantar2022gptq} highlights that LLMs can still achieve a desirable accuracy with only 4-bit precision quantization. 
%In order to take the activations into account for quantizing the weights, 
Furthermore, Lin {\em et al.}~\shortcite{lin2023awq} find that weights do not carry equal importance for the performance, and accordingly, the quantization error can be significantly reduced by maintaining only {1\%} of salient weights in full-precision. They propose Activation-aware Quantization (AWQ) method, which retains the weights corresponding to large activations in full-precision.
%in which the weights retained in full-precision are those corresponding to larger activation magnitudes.
In order to address the significant quantization error resulted from the outliers in activations distribution,
%One of the main challenges in quantizing activations is that the outliers in activations distribution can impose a significant error.% in the quantization. 
%Providing that insight, 
Lee {\em et al.}~\shortcite{lee2023owq} propose a mixed-precision quantization approach, namely, outlier-aware weight quantization (OWQ), which applies higher precision to the weights associated with outlier activations.
\paragraph{Pruning.}
Pruning refers to reducing the size of LLMs by removing redundant parameters that are less crucial for the models. Pruning can help optimize the model for deployment and make the model more efficient in terms of computation complexity and memory usage. Accordingly, pruning can be considered as an approach to enable a language model to support longer sequence length, while maintaining the desirable complexity and performance. In general, pruning a model can be categorized into structured and unstructured pruning. 

Structured pruning aims at removing higher-granularity structures, such as entire neurons, layers, or rows/columns of weight matrices, which can result in a model that retains its original structure but with fewer number of parameters.
%and smaller size. %Hence, structured pruning can make the models more hardware-friendly for deployment. 
LLM-Pruner~\cite{ma2023llm} is a structural task-agnostic pruning approach that selectively removes non-critical connection structures considering both first-order information and an approximated Hessian information  gradient information. 
%Different pruning approaches based on Low-rank Adaptation (LoRA) \cite{hu2021lora} have also been studied in the literature. LoRAPrun \cite{loraprune} is a framework that utilizes a pruning metric that is obtained through the weights and gradients of LoRA, rather than the gradients of pre-trained weights. 
%Then it removes the channels and head in a structured pruning structure. 
%LoRAPrune outperforms LLMPruner in efficiency at a 50{\%} compression rate. LoRAShear \cite{chen2023lorashear} is then proposed to address the significant performance degradation in LoRAPrune, in a limited-resource environment. It utilizes a sparse optimizer called LoRA Half-Space Projected Gradient (LHSPG) to perform a progressive structured pruning and transfer the knowledge. 
As an alternative, Sheared LLaMA \cite{xia2023sheared} utilizes a two-stage approach for pruning an LLM. In the first stage, it exploits targeted structured pruning to prune a large model to a targeted shape by pruning layers, heads, and intermediate connections. In the second stage, the batches of data are loaded dynamically and the model structure is modified in each training iteration based on losses in various domains. As a result, Sheared LLaMA achieves a compressed model that can outperform the LLMs, with the same size but trained from scratch.  %However, Sheared LLaMA has a stringent requirements on the computation and data resources. 

Unstructured pruning involves with pruning individual parameters of a model independently based on their magnitudes or importance, resulting in an irregular sparse structure. Due to the irregularity in the structure and in the memory access patterns, unstructured pruning hinders the efficiency gain that might be achieved through structured pruning, and it requires specialized software and/or hardware for efficient deployment. SparseGPT~\cite{frantar2023sparsegpt} %is an unstructured pruning approach that can 
compresses LLMs with billions of parameter by as much as 60\%, almost without affecting the performance of the models. However, SparseGPT heavily relies on weight updates. In order to address this issue, Sun {\em et al.}~\shortcite{sun2023simple} propose Wanda that prunes the weights according to novel criterion, which is mainly based on product value of the weights and their input activations. 

%\subsection{Computation Compression}
% transition paragraph is needed here!
%%%%%%%%%%%%%%%% Xiangyu %%%%%%%%%%%%%%%%%
\paragraph{Multi-query and Group Attention.}
While multi-head attention has demonstrated its effectiveness in characterizing the correlations among tokens, it suffers from the incremental memory bandwidth cost and longer latency during inference due to repeatedly loading the large $\mathbf{KV}$ tensors as the input sequence length increases.
Multi-query attention (MQA)~\cite{shazeer2019fast} is one of the approaches that address the aforementioned issue. In particular,
%the incremental memory bandwidth cost and longer latency issues due to repeatedly loading the large $\mathbf{KV}$ tensors as the input sequence length increases. 
MQA essentially reuses the same $\mathbf{KV}$ tensors across all attention heads of each query to reduce the memory bandwidth requirements during decoding and thus, allows longer sequences and faster decoding. Given its demonstrated performance with minor quality degradation, MQA has been adopted in several works. Google~\cite{chowdhery2023palm} trains a LLM named Pathways Language Model (PaLM) with the adoption of MQA to improve decoding speed and later PaLM-2~\cite{anil2023palm} is released with improved computation efficiency. 
%Reiner {\em et al.}~\shortcite{pope2023efficiently} 
Pope {\em et al.}~\shortcite{pope2023efficiently} propose a partition-optimized model that enables up to $32\times$ larger context lengths with the help of MQA on LLMs. de Jong {\em et al.}~\shortcite{de-jong-etal-2023-fido} adopts MQA to reduce the cross-attention computation at the decoders in Fusion-in-Decoder models with faster inference. More recently, Li {\em et al.}~\shortcite{li2023starcoder} introduce StarCoder, a Code LLM, with fast large-batch inference enabled by MQA. The shared $\mathbf{KV}$ tensors idea in MQA also inspired the emergence of other attention computation schemes. A grouped-query attention (GQA)~\cite{ainslie-etal-2023-gqa} mechanism is proposed to trade-off performance degradation and speed by sharing a subset of $\mathbf {KV}$ tensors. 
%SparQ attention proposed in \cite{ribar2023sparq} reduces memory bandwidth requirement by selecting top-$K$ largest components in the $\mathbf{KV}$ tensors for attention calculation. 

%%%%%%%%%%%%%%%% Parsa %%%%%%%%%%%%%%%%%
\section{Hardware-aware Transformers}
A viable solution to challenges posed by long sequence in LLMs involves adapting algorithms to be hardware-aware, enhancing efficiency and enabling the processing of longer sequences. Our exploration encompasses a spectrum of innovations, each tailored to address distinct aspects of IO-awareness, resource management, multi-device distributed attention, and memory management. 
\paragraph{IO-awareness and Resource Management.} A critical concern in deep neural network models like transformers is the constant need for Read/Write operations from/to memory. FlashAttention \cite{dao2022flashattention} %\cite{dao2022flashattention,dao2023flashattention}
addresses this challenge by making attention algorithms IO-aware, effectively managing reads and writes between different levels of GPU memory. This approach capitalizes on the insight that the softmax matrix in attention can be computed without materializing the entire matrix, utilizing tiling techniques. FlashAttention introduces parallelization over sequence length, processing different portions (blocks) of the sequence to compute attention in a more manageable block-wise operation within fast memory Static Random Access Memory (SRAM) in GPUs. Moreover, FlashAttention highlights the efficiency gains of recomputing attention during the backward pass of optimization. It utilizes blocks already present in SRAM instead of storing attention results in high bandwidth memory (HBM) and transferring them to SRAM again. Building on FlashAttention foundations, Block-wise Parallel Transformer (BPT) \cite{liu2023blockwise} fuses the feedforward layer with self-attention to further minimize IO usage, enabling the model to handle sequences up to four times longer than FlashAttention.

This IO-aware approach is not exclusive to attention-based transformers; similar techniques have been applied to expedite SSMs. SSMs, emerging as alternatives to transformers due to linear scalability and convolution implementation feasibility, present challenges in convolution-dominated computation time during training. FlashConv \cite{fu2023hungry} addresses this by leveraging the Cooley-Tukey decomposition of the FFT into a series of diagonal matrix multiplication, to take advantage of fast tensor cores. To accommodate longer sequences, FlashConv utilizes the recurrent properties of SSMs, allowing convolution to occur in different portions sequentially. Mamba~\cite{gu2023mamba} further enhances SSMs through innovative techniques such as kernel fusion, parallel scan, and recomputation, leveraging modern accelerators like GPUs for efficient memory hierarchy utilization.

\paragraph{Multi-device Distributed Attention.} Both FlashAttention and BPT leverage distinct streaming multiprocessors in GPUs for parallel processing of different blocks. However, the limited size of SRAM imposes constraints on sequence length. Encouragingly, this concept can be expanded to accommodate very long sequences by distributing attention computation across multiple GPUs, as proposed by Ring Attention \cite{liu2023ring}. This innovative approach enables block-wise self-attention computation, seamlessly overlapping communication of key-value blocks among devices with the computation of each block on devices. As a result, it facilitates the processing of sequences several times longer than BPT, showcasing scalability across the device count. Another example of distributing attention computation across multiple devices is demonstrated by LongNet \cite{ding2023longnet}. LongNet possesses the capability to compute multiple attentions, each with a distinct dilated sparsity pattern. These computations operate independently and can be distributed over multiple devices, with each device corresponding to a single dilated pattern. This collective approach facilitates the processing of longer sequences.

\paragraph{Memory Management.} 
Effective memory management is vital in LLMs, especially during the autoregressive inference phase. The sequential generation of tokens, repeated for each request, leads to a memory-bound workload, limiting GPU utilization and serving throughput. To enhance throughput, batching multiple requests requires efficient memory management, specifically for Key-Value ($\mathbf{KV}$) caches. The dynamic nature of $\mathbf{KV}$ cache growth and its unpredictable lifetime necessitate adaptive strategies for optimal memory utilization in varying context lengths. 

PagedAttention \cite{kwon2023efficient} employs a virtual memory-inspired strategy during the inference phase to tackle the memory-bound challenges inherent in sequential generation. By segmenting $\mathbf{KV}$ caches into blocks, this approach achieves flexible memory management, effectively mitigating both internal and external fragmentation issues. In the pursuit of attention acceleration during inference, Flash-Decoding \cite{dao2023flash} builds upon FlashAttention principles. Introducing a new parallelization dimension for keys/values sequence length, it ensures optimal GPU utilization even with small batch sizes and large context lengths. This approach proves instrumental in achieving up to $8\times$ faster generation for very long sequences. Additionally, other methods enhance memory management efficiency. %PO: FlashDecoding++ section (next sentence) can be removed to get some space
FlashDecoding++~\cite{hong2023flashdecoding++}, for instance, goes one step further by eliminating the need for synchronization in handling partial softmax computations, effectively addressing a limitation observed in prior works.

Another notable memory management technique is LLM-in-Flash ~\cite{alizadeh2023llm}, which leverages the larger size of flash memory compared to Dynamic Random Access Memory (DRAM). This method runs an LLM during inference efficiently by storing model parameters in flash memory and bringing them to DRAM on demand. To balance the lower bandwidth of Flash memory, the paper introduces an inference cost model considering flash memory characteristics. The technique incorporates sparsity awareness in feedforward layers and context-adaptive loading of the model. Although not specifically used to increase sequence length, this method has the potential to load a model up to twice the size of the available DRAM. This capacity could be harnessed to handle longer sequences while ensuring practical data transfer between DRAM and flash memory. 

\section{Conclusion and Future Directions}
In this survey, a systematic review of different approaches for efficiently extending the sequence length in LLMs is provided. We start with the motivation of the work and the necessity of handling long sequences by LLMs. We then discuss the methods that encompass architectural changes, such as positional encoding modification, and attention mechanism modification,% and model compression strategies, 
designed to improve %the handling of longer sequences 
long sequences handling without significantly increasing the computational cost. We further explore the methods that can be applied to different phases, such as training, fine-tuning and inference, which efficiently improve the LLM's capability of processing long sequences. %and optimization strategies that have been crucial in equipping LLMs to efficiently process extended sequences. 
These techniques not only address the immediate challenges posed by sequence length limitations but also pave the way for more complex and contextually aware LLMs.

%However, the journey towards fully optimizing LLMs for long-sequence processing is far from complete. The field stands on the brink of numerous possibilities and uncharted territories. Future research could concentrate on developing techniques that not only enable the LLMs to process longer sequences but also further reduce the computational and memory demands, which will be crucial. Moreover, the development of more refined attention mechanisms that can efficiently handle long sequences without losing contextual understanding represents a promising area of research. There is also a need for comprehensive benchmarking and evaluation frameworks that could accurately examine the capabilities of LLMs in handling extended contexts. This includes creating datasets that specifically challenge and test the long-sequence processing capabilities of LLMs.

Despite these advancements, challenges related to computational cost, model interpretability, and the ability to integrate external knowledge remain prevalent. The trade-offs between model complexity, processing speed, and accuracy continue to be a pivotal consideration in the design and implementation of LLMs for long sequences. Future research could focus on further optimizing the architecture of LLMs to enhance their efficiency in processing long sequences. Innovations could involve %the development of 
developing %more sophisticated 
attention mechanisms or network structures that can handle long sequences more effectively while not increasing the computational cost. In addition, integrating LLMs with external knowledge could improve their ability in understanding and generating longer coherent and contextually accurate sequences. Exploring methods for effective knowledge integration and retrieval during the language generation process would be beneficial, too. Moreover, new training methodologies can be investigated to improve the ability of the model to understand and retain information over longer sequences. Techniques such as curriculum learning, where models are gradually exposed to increasingly longer sequences during training, could be one direction to explore. Last but not least, there is also a need for comprehensive benchmarking and evaluation frameworks that could accurately examine the capabilities of LLMs in handling long sequences. This includes creating datasets that specifically challenge the long-context processing capabilities of LLMs.
\appendix

%\section*{Ethical Statement}
%This is a review paper that analyses the existing techniques on LLMs dealing with longer sequences. We do not see any ethics issues here in this paper.

%\section*{Acknowledgments}

%\newpage
%% The file named.bst is a bibliography style file for BibTeX 0.99c
\bibliographystyle{named}
\bibliography{ijcai24_full}

\begin{thebibliography}{}

\bibitem[\protect\citeauthoryear{Ainslie \bgroup \em et al.\egroup }{2023}]{ainslie-etal-2023-gqa}
Joshua Ainslie, James Lee-Thorp, Michiel de~Jong, Yury Zemlyanskiy, Federico Lebron, and Sumit Sanghai.
\newblock {GQA}: Training generalized multi-query transformer models from multi-head checkpoints.
\newblock In {\em Proceedings of the 2023 Conference on Empirical Methods in Natural Language Processing}, pages 4895--4901, Singapore, December 2023. Association for Computational Linguistics.

\bibitem[\protect\citeauthoryear{Alizadeh \bgroup \em et al.\egroup }{2023}]{alizadeh2023llm}
Keivan Alizadeh, Iman Mirzadeh, Dmitry Belenko, Karen Khatamifard, Minsik Cho, Carlo~C Del~Mundo, Mohammad Rastegari, and Mehrdad Farajtabar.
\newblock {LLM} in a flash: Efficient large language model inference with limited memory.
\newblock {\em arXiv preprint arXiv:2312.11514}, 2023.

\bibitem[\protect\citeauthoryear{Anil \bgroup \em et al.\egroup }{2023}]{anil2023palm}
Rohan Anil, Andrew~M Dai, Orhan Firat, Melvin Johnson, Dmitry Lepikhin, Alexandre Passos, Siamak Shakeri, Emanuel Taropa, Paige Bailey, Zhifeng Chen, et~al.
\newblock Pa{LM} 2 technical report.
\newblock {\em arXiv preprint arXiv:2305.10403}, 2023.

\bibitem[\protect\citeauthoryear{Beltagy \bgroup \em et al.\egroup }{2020}]{beltagy2020longformer}
Iz~Beltagy, Matthew~E Peters, and Arman Cohan.
\newblock Longformer: The long-document transformer.
\newblock {\em arXiv preprint arXiv:2004.05150}, 2020.

\bibitem[\protect\citeauthoryear{Brown \bgroup \em et al.\egroup }{2020}]{NEURIPS2020_1457c0d6}
Tom~B. Brown, Benjamin Mann, Nick Ryder, Melanie Subbiah, Jared Kaplan, Prafulla Dhariwal, Arvind Neelakantan, Pranav Shyam, Girish Sastry, Amanda Askell, Sandhini Agarwal, Ariel Herbert-Voss, Gretchen Krueger, Tom Henighan, Rewon Child, Aditya Ramesh, Daniel~M. Ziegler, Jeffrey Wu, Clemens Winter, Christopher Hesse, Mark Chen, Eric Sigler, Mateusz Litwin, Scott Gray, Benjamin Chess, Jack Clark, Christopher Berner, Sam McCandlish, Alec Radford, Ilya Sutskever, and Dario Amodei.
\newblock Language models are few-shot learners.
\newblock In {\em Proceedings of the 34th International Conference on Neural Information Processing Systems}, NIPS '20, 2020.

\bibitem[\protect\citeauthoryear{Chen \bgroup \em et al.\egroup }{2021}]{chen-etal-2021-simple}
Pu-Chin Chen, Henry Tsai, Srinadh Bhojanapalli, Hyung~Won Chung, Yin-Wen Chang, and Chun-Sung Ferng.
\newblock A simple and effective positional encoding for transformers.
\newblock In {\em Proceedings of the 2021 Conference on Empirical Methods in Natural Language Processing}, pages 2974--2988, Online and Punta Cana, Dominican Republic, November 2021. Association for Computational Linguistics.

\bibitem[\protect\citeauthoryear{Chen \bgroup \em et al.\egroup }{2023}]{chen2023extending}
Shouyuan Chen, Sherman Wong, Liangjian Chen, and Yuandong Tian.
\newblock Extending context window of large language models via positional interpolation.
\newblock {\em arXiv preprint arXiv:2306.15595}, 2023.

\bibitem[\protect\citeauthoryear{Chen \bgroup \em et al.\egroup }{2024}]{chen2024clex}
Guanzheng Chen, Xin Li, Zaiqiao Meng, Shangsong Liang, and Lidong Bing.
\newblock {CLEX}: Continuous length extrapolation for large language models.
\newblock In {\em The Twelfth International Conference on Learning Representations}, 2024.

\bibitem[\protect\citeauthoryear{Choromanski \bgroup \em et al.\egroup }{2021}]{choromanski2021rethinking}
Krzysztof~Marcin Choromanski, Valerii Likhosherstov, David Dohan, Xingyou Song, Andreea Gane, Tamas Sarlos, Peter Hawkins, Jared~Quincy Davis, Afroz Mohiuddin, Lukasz Kaiser, David~Benjamin Belanger, Lucy~J Colwell, and Adrian Weller.
\newblock Rethinking attention with performers.
\newblock In {\em International Conference on Learning Representations}, 2021.

\bibitem[\protect\citeauthoryear{Chowdhery \bgroup \em et al.\egroup }{2024}]{chowdhery2023palm}
Aakanksha Chowdhery, Sharan Narang, Jacob Devlin, Maarten Bosma, Gaurav Mishra, Adam Roberts, Paul Barham, Hyung~Won Chung, Charles Sutton, Sebastian Gehrmann, Parker Schuh, Kensen Shi, Sashank Tsvyashchenko, Joshua Maynez, Abhishek Rao, Parker Barnes, Yi~Tay, Noam Shazeer, Vinodkumar Prabhakaran, Emily Reif, Nan Du, Ben Hutchinson, Reiner Pope, James Bradbury, Jacob Austin, Michael Isard, Guy Gur-Ari, Pengcheng Yin, Toju Duke, Anselm Levskaya, Sanjay Ghemawat, Sunipa Dev, Henryk Michalewski, Xavier Garcia, Vedant Misra, Kevin Robinson, Liam Fedus, Denny Zhou, Daphne Ippolito, David Luan, Hyeontaek Lim, Barret Zoph, Alexander Spiridonov, Ryan Sepassi, David Dohan, Shivani Agrawal, Mark Omernick, Andrew~M. Dai, Thanumalayan~Sankaranarayana Pillai, Marie Pellat, Aitor Lewkowycz, Erica Moreira, Rewon Child, Oleksandr Polozov, Katherine Lee, Zongwei Zhou, Xuezhi Wang, Brennan Saeta, Mark Diaz, Orhan Firat, Michele Catasta, Jason Wei, Kathy Meier-Hellstern, Douglas Eck, Jeff Dean, Slav Petrov, and Noah Fiedel.
\newblock {PaLM}: scaling language modeling with pathways.
\newblock {\em Journal of Machine Learning Research}, 24(1), March 2024.

\bibitem[\protect\citeauthoryear{Dao \bgroup \em et al.\egroup }{2022}]{dao2022flashattention}
Tri Dao, Dan Fu, Stefano Ermon, Atri Rudra, and Christopher R\'{e}.
\newblock Flashattention: Fast and memory-efficient exact attention with io-awareness.
\newblock In {\em Advances in Neural Information Processing Systems}, volume~35, pages 16344--16359, 2022.

\bibitem[\protect\citeauthoryear{Dao \bgroup \em et al.\egroup }{2023}]{dao2023flash}
Tri Dao, Daniel Haziza, Francisco Massa, and Grigory Sizov.
\newblock Flash-{D}ecoding for long-context inference, 2023.

\bibitem[\protect\citeauthoryear{de Jong \bgroup \em et al.\egroup }{2023}]{de-jong-etal-2023-fido}
Michiel de~Jong, Yury Zemlyanskiy, Joshua Ainslie, Nicholas FitzGerald, Sumit Sanghai, Fei Sha, and William Cohen.
\newblock {F}i{DO}: Fusion-in-decoder optimized for stronger performance and faster inference.
\newblock In {\em Findings of the Association for Computational Linguistics: ACL 2023}, pages 11534--11547, Toronto, Canada, July 2023. Association for Computational Linguistics.

\bibitem[\protect\citeauthoryear{Dettmers \bgroup \em et al.\egroup }{2022}]{NEURIPS2022_c3ba4962}
Tim Dettmers, Mike Lewis, Younes Belkada, and Luke Zettlemoyer.
\newblock {LLM}.int8(): 8-bit matrix multiplication for transformers at scale.
\newblock In {\em Advances in Neural Information Processing Systems}, volume~35, pages 30318--30332, 2022.

\bibitem[\protect\citeauthoryear{Ding \bgroup \em et al.\egroup }{2023}]{ding2023longnet}
Jiayu Ding, Shuming Ma, Li~Dong, Xingxing Zhang, Shaohan Huang, Wenhui Wang, Nanning Zheng, and Furu Wei.
\newblock Long{N}et: Scaling transformers to 1,000,000,000 tokens.
\newblock {\em arXiv preprint arXiv:2307.02486}, 2023.

\bibitem[\protect\citeauthoryear{Frantar and Alistarh}{2023}]{frantar2023sparsegpt}
Elias Frantar and Dan Alistarh.
\newblock {S}parse{GPT}: Massive language models can be accurately pruned in one-shot.
\newblock In {\em Proceedings of the 40th International Conference on Machine Learning}, volume 202 of {\em Proceedings of Machine Learning Research}, pages 10323--10337. PMLR, 23--29 Jul 2023.

\bibitem[\protect\citeauthoryear{Frantar \bgroup \em et al.\egroup }{2023}]{frantar2023optq}
Elias Frantar, Saleh Ashkboos, Torsten Hoefler, and Dan Alistarh.
\newblock {OPTQ}: Accurate quantization for generative pre-trained transformers.
\newblock In {\em The Eleventh International Conference on Learning Representations}, 2023.

\bibitem[\protect\citeauthoryear{Fu \bgroup \em et al.\egroup }{2023}]{fu2023hungry}
Daniel~Y Fu, Tri Dao, Khaled~Kamal Saab, Armin~W Thomas, Atri Rudra, and Christopher Re.
\newblock Hungry {H}ungry {H}ippos: Towards language modeling with state space models.
\newblock In {\em The Eleventh International Conference on Learning Representations}, 2023.

\bibitem[\protect\citeauthoryear{Gholami \bgroup \em et al.\egroup }{2022}]{gholami2022survey}
Amir Gholami, Sehoon Kim, Zhen Dong, Zhewei Yao, Michael~W Mahoney, and Kurt Keutzer.
\newblock A survey of quantization methods for efficient neural network inference.
\newblock In {\em Low-Power Computer Vision}, pages 291--326. Chapman and Hall/CRC, 2022.

\bibitem[\protect\citeauthoryear{Gu and Dao}{2023}]{gu2023mamba}
Albert Gu and Tri Dao.
\newblock Mamba: Linear-time sequence modeling with selective state spaces.
\newblock {\em arXiv preprint arXiv:2312.00752}, 2023.

\bibitem[\protect\citeauthoryear{Gu \bgroup \em et al.\egroup }{2020}]{NEURIPS2020_102f0bb6}
Albert Gu, Tri Dao, Stefano Ermon, Atri Rudra, and Christopher R\'{e}.
\newblock Hippo: Recurrent memory with optimal polynomial projections.
\newblock In {\em Advances in Neural Information Processing Systems}, volume~33, pages 1474--1487, 2020.

\bibitem[\protect\citeauthoryear{Gu \bgroup \em et al.\egroup }{2022}]{gu2022efficiently}
Albert Gu, Karan Goel, and Christopher Re.
\newblock Efficiently modeling long sequences with structured state spaces.
\newblock In {\em International Conference on Learning Representations}, 2022.

\bibitem[\protect\citeauthoryear{Hao \bgroup \em et al.\egroup }{2022}]{hao2022structured}
Yaru Hao, Yutao Sun, Li~Dong, Zhixiong Han, Yuxian Gu, and Furu Wei.
\newblock Structured prompting: Scaling in-context learning to 1,000 examples.
\newblock {\em arXiv preprint arXiv:2212.06713}, 2022.

\bibitem[\protect\citeauthoryear{Hong \bgroup \em et al.\egroup }{2023}]{hong2023flashdecoding++}
Ke~Hong, Guohao Dai, Jiaming Xu, Qiuli Mao, Xiuhong Li, Jun Liu, Kangdi Chen, Hanyu Dong, and Yu~Wang.
\newblock {FlashDecoding}++: Faster large language model inference on gpus.
\newblock {\em arXiv preprint arXiv:2311.01282}, 2023.

\bibitem[\protect\citeauthoryear{Huang \bgroup \em et al.\egroup }{2023}]{huang2023advancing}
Yunpeng Huang, Jingwei Xu, Zixu Jiang, Junyu Lai, Zenan Li, Yuan Yao, Taolue Chen, Lijuan Yang, Zhou Xin, and Xiaoxing Ma.
\newblock Advancing transformer architecture in long-context large language models: A comprehensive survey.
\newblock {\em arXiv preprint arXiv:2311.12351}, 2023.

\bibitem[\protect\citeauthoryear{Jiang \bgroup \em et al.\egroup }{2023a}]{jiang-etal-2023-llmlingua}
Huiqiang Jiang, Qianhui Wu, Chin-Yew Lin, Yuqing Yang, and Lili Qiu.
\newblock {LLML}ingua: Compressing prompts for accelerated inference of large language models.
\newblock In {\em Proceedings of the 2023 Conference on Empirical Methods in Natural Language Processing}, pages 13358--13376, Singapore, December 2023. Association for Computational Linguistics.

\bibitem[\protect\citeauthoryear{Jiang \bgroup \em et al.\egroup }{2023b}]{jiang2023longllmlingua}
Huiqiang Jiang, Qianhui Wu, Xufang Luo, Dongsheng Li, Chin-Yew Lin, Yuqing Yang, and Lili Qiu.
\newblock Long{LLML}ingua: Accelerating and enhancing {LLMs} in long context scenarios via prompt compression.
\newblock {\em arXiv preprint arXiv:2310.06839}, 2023.

\bibitem[\protect\citeauthoryear{Kitaev \bgroup \em et al.\egroup }{2020}]{Kitaev2020Reformer}
Nikita Kitaev, Lukasz Kaiser, and Anselm Levskaya.
\newblock Reformer: The efficient transformer.
\newblock In {\em International Conference on Learning Representations}, 2020.

\bibitem[\protect\citeauthoryear{Koohpayegani and Pirsiavash}{2024}]{10483688}
Soroush~Abbasi Koohpayegani and Hamed Pirsiavash.
\newblock Sim{A}: Simple softmax-free attention for vision transformers.
\newblock In {\em 2024 IEEE/CVF Winter Conference on Applications of Computer Vision (WACV)}, pages 2595--2605, 2024.

\bibitem[\protect\citeauthoryear{Kwan \bgroup \em et al.\egroup }{2023}]{kwan2023m4le}
Wai-Chung Kwan, Xingshan Zeng, Yufei Wang, Yusen Sun, Liangyou Li, Lifeng Shang, Qun Liu, and Kam-Fai Wong.
\newblock {M4LE}: A multi-ability multi-range multi-task multi-domain long-context evaluation benchmark for large language models.
\newblock {\em arXiv preprint arXiv:2310.19240}, 2023.

\bibitem[\protect\citeauthoryear{Kwon \bgroup \em et al.\egroup }{2023}]{kwon2023efficient}
Woosuk Kwon, Zhuohan Li, Siyuan Zhuang, Ying Sheng, Lianmin Zheng, Cody~Hao Yu, Joseph Gonzalez, Hao Zhang, and Ion Stoica.
\newblock Efficient memory management for large language model serving with pagedattention.
\newblock In {\em Proceedings of the 29th Symposium on Operating Systems Principles}, SOSP '23, page 611–626, New York, NY, USA, 2023. Association for Computing Machinery.

\bibitem[\protect\citeauthoryear{Lee \bgroup \em et al.\egroup }{2024}]{lee2023owq}
Changhun Lee, Jungyu Jin, Taesu Kim, Hyungjun Kim, and Eunhyeok Park.
\newblock {OWQ}: Outlier-aware weight quantization for efficient fine-tuning and inference of large language models.
\newblock In {\em Proceedings of the AAAI Conference on Artificial Intelligence}, volume~38, pages 13355--13364, 2024.

\bibitem[\protect\citeauthoryear{Li \bgroup \em et al.\egroup }{2019}]{NEURIPS2019_6775a063}
Shiyang Li, Xiaoyong Jin, Yao Xuan, Xiyou Zhou, Wenhu Chen, Yu-Xiang Wang, and Xifeng Yan.
\newblock Enhancing the locality and breaking the memory bottleneck of transformer on time series forecasting.
\newblock In {\em Advances in Neural Information Processing Systems}, volume~32, 2019.

\bibitem[\protect\citeauthoryear{Li \bgroup \em et al.\egroup }{2023a}]{li2023starcoder}
Raymond Li, Loubna~Ben allal, Yangtian Zi, Niklas Muennighoff, Denis Kocetkov, Chenghao Mou, Marc Marone, Christopher Akiki, Jia LI, Jenny Chim, Qian Liu, Evgenii Zheltonozhskii, Terry~Yue Zhuo, Thomas Wang, Olivier Dehaene, Joel Lamy-Poirier, Joao Monteiro, Nicolas Gontier, Ming-Ho Yee, Logesh~Kumar Umapathi, Jian Zhu, Ben Lipkin, Muhtasham Oblokulov, Zhiruo Wang, Rudra Murthy, Jason~T Stillerman, Siva~Sankalp Patel, Dmitry Abulkhanov, Marco Zocca, Manan Dey, Zhihan Zhang, Urvashi Bhattacharyya, Wenhao Yu, Sasha Luccioni, Paulo Villegas, Fedor Zhdanov, Tony Lee, Nadav Timor, Jennifer Ding, Claire~S Schlesinger, Hailey Schoelkopf, Jan Ebert, Tri Dao, Mayank Mishra, Alex Gu, Carolyn~Jane Anderson, Brendan Dolan-Gavitt, Danish Contractor, Siva Reddy, Daniel Fried, Dzmitry Bahdanau, Yacine Jernite, Carlos~Mu{\~n}oz Ferrandis, Sean Hughes, Thomas Wolf, Arjun Guha, Leandro~Von Werra, and Harm de~Vries.
\newblock Star{C}oder: may the source be with you!
\newblock {\em Transactions on Machine Learning Research}, 2023.

\bibitem[\protect\citeauthoryear{Li \bgroup \em et al.\egroup }{2023b}]{li-etal-2023-compressing}
Yucheng Li, Bo~Dong, Frank Guerin, and Chenghua Lin.
\newblock Compressing context to enhance inference efficiency of large language models.
\newblock In {\em Proceedings of the 2023 Conference on Empirical Methods in Natural Language Processing}, pages 6342--6353, Singapore, December 2023. Association for Computational Linguistics.

\bibitem[\protect\citeauthoryear{Lin \bgroup \em et al.\egroup }{2024}]{lin2023awq}
Ji~Lin, Jiaming Tang, Haotian Tang, Shang Yang, Wei-Ming Chen, Wei-Chen Wang, Guangxuan Xiao, Xingyu Dang, Chuang Gan, and Song Han.
\newblock {AWQ}: Activation-aware weight quantization for llm compression and acceleration.
\newblock In {\em The Seventh Annual Conference on Machine Learning and Systems}, 2024.

\bibitem[\protect\citeauthoryear{Liu and Abbeel}{2023}]{liu2023blockwise}
Hao Liu and Pieter Abbeel.
\newblock Blockwise parallel transformers for large context models.
\newblock In {\em Advances in Neural Information Processing Systems}, volume~36, pages 8828--8844, 2023.

\bibitem[\protect\citeauthoryear{Liu \bgroup \em et al.\egroup }{2023a}]{liu2023ring}
Hao Liu, Matei Zaharia, and Pieter Abbeel.
\newblock Ring attention with blockwise transformers for near-infinite context.
\newblock In {\em NeurIPS 2023 Foundation Models for Decision Making Workshop}, 2023.

\bibitem[\protect\citeauthoryear{Liu \bgroup \em et al.\egroup }{2023b}]{liu2023llm}
Zechun Liu, Barlas Oguz, Changsheng Zhao, Ernie Chang, Pierre Stock, Yashar Mehdad, Yangyang Shi, Raghuraman Krishnamoorthi, and Vikas Chandra.
\newblock {LLM-QAT}: Data-free quantization aware training for large language models.
\newblock {\em arXiv preprint arXiv:2305.17888}, 2023.

\bibitem[\protect\citeauthoryear{Lu \bgroup \em et al.\egroup }{2021}]{NEURIPS2021_b1d10e7b}
Jiachen Lu, Jinghan Yao, Junge Zhang, Xiatian Zhu, Hang Xu, Weiguo Gao, Chunjing XU, Tao Xiang, and Li~Zhang.
\newblock {SOFT}: Softmax-free transformer with linear complexity.
\newblock In {\em Advances in Neural Information Processing Systems}, volume~34, pages 21297--21309, 2021.

\bibitem[\protect\citeauthoryear{Ma \bgroup \em et al.\egroup }{2023a}]{ma2023llm}
Xinyin Ma, Gongfan Fang, and Xinchao Wang.
\newblock {LLM-Pruner}: On the structural pruning of large language models.
\newblock In {\em Advances in Neural Information Processing Systems}, volume~36, pages 21702--21720, 2023.

\bibitem[\protect\citeauthoryear{Ma \bgroup \em et al.\egroup }{2023b}]{ma2023mega}
Xuezhe Ma, Chunting Zhou, Xiang Kong, Junxian He, Liangke Gui, Graham Neubig, Jonathan May, and Luke Zettlemoyer.
\newblock {MEGA}: Moving average equipped gated attention.
\newblock In {\em The Eleventh International Conference on Learning Representations}, 2023.

\bibitem[\protect\citeauthoryear{Naveed \bgroup \em et al.\egroup }{2023}]{naveed2023comprehensive}
Humza Naveed, Asad~Ullah Khan, Shi Qiu, Muhammad Saqib, Saeed Anwar, Muhammad Usman, Nick Barnes, and Ajmal Mian.
\newblock A comprehensive overview of large language models.
\newblock {\em arXiv preprint arXiv:2307.06435}, 2023.

\bibitem[\protect\citeauthoryear{Orvieto \bgroup \em et al.\egroup }{2023}]{orvieto2023resurrecting}
Antonio Orvieto, Samuel~L Smith, Albert Gu, Anushan Fernando, Caglar Gulcehre, Razvan Pascanu, and Soham De.
\newblock Resurrecting recurrent neural networks for long sequences.
\newblock In {\em Proceedings of the 40th International Conference on Machine Learning}, ICML'23. JMLR.org, 2023.

\bibitem[\protect\citeauthoryear{Packer \bgroup \em et al.\egroup }{2023}]{packer2023memgpt}
Charles Packer, Vivian Fang, Shishir~G Patil, Kevin Lin, Sarah Wooders, and Joseph~E Gonzalez.
\newblock Mem{GPT}: Towards {LLMs} as operating systems.
\newblock {\em arXiv preprint arXiv:2310.08560}, 2023.

\bibitem[\protect\citeauthoryear{Peng \bgroup \em et al.\egroup }{2023a}]{peng-etal-2023-rwkv}
Bo~Peng, Eric Alcaide, Quentin Anthony, Alon Albalak, Samuel Arcadinho, Stella Biderman, Huanqi Cao, Xin Cheng, Michael Chung, Leon Derczynski, Xingjian Du, Matteo Grella, Kranthi Gv, Xuzheng He, Haowen Hou, Przemyslaw Kazienko, Jan Kocon, Jiaming Kong, Bart{\l}omiej Koptyra, Hayden Lau, Jiaju Lin, Krishna Sri~Ipsit Mantri, Ferdinand Mom, Atsushi Saito, Guangyu Song, Xiangru Tang, Johan Wind, Stanis{\l}aw Wo{\'z}niak, Zhenyuan Zhang, Qinghua Zhou, Jian Zhu, and Rui-Jie Zhu.
\newblock {RWKV}: Reinventing {RNN}s for the transformer era.
\newblock In {\em Findings of the Association for Computational Linguistics: EMNLP 2023}, pages 14048--14077, Singapore, December 2023. Association for Computational Linguistics.

\bibitem[\protect\citeauthoryear{Peng \bgroup \em et al.\egroup }{2023b}]{peng2023yarn}
Bowen Peng, Jeffrey Quesnelle, Honglu Fan, and Enrico Shippole.
\newblock {YaRN}: Efficient context window extension of large language models.
\newblock In {\em The Eleventh International Conference on Learning Representations}, 2023.

\bibitem[\protect\citeauthoryear{Poli \bgroup \em et al.\egroup }{2023}]{poli2023hyena}
Michael Poli, Stefano Massaroli, Eric Nguyen, Daniel~Y. Fu, Tri Dao, Stephen Baccus, Yoshua Bengio, Stefano Ermon, and Christopher R\'{e}.
\newblock Hyena {H}ierarchy: towards larger convolutional language models.
\newblock In {\em Proceedings of the 40th International Conference on Machine Learning}, ICML'23. JMLR.org, 2023.

\bibitem[\protect\citeauthoryear{Pope \bgroup \em et al.\egroup }{2023}]{pope2023efficiently}
Reiner Pope, Sholto Douglas, Aakanksha Chowdhery, Jacob Devlin, James Bradbury, Jonathan Heek, Kefan Xiao, Shivani Agrawal, and Jeff Dean.
\newblock Efficiently scaling transformer inference.
\newblock {\em Proceedings of Machine Learning and Systems}, 5, 2023.

\bibitem[\protect\citeauthoryear{Press \bgroup \em et al.\egroup }{2022}]{press2022train}
Ofir Press, Noah Smith, and Mike Lewis.
\newblock Train short, test long: Attention with linear biases enables input length extrapolation.
\newblock In {\em International Conference on Learning Representations}, 2022.

\bibitem[\protect\citeauthoryear{Qin \bgroup \em et al.\egroup }{2022}]{zhen2022cosformer}
Zhen Qin, Weixuan Sun, Hui Deng, Dongxu Li, Yunshen Wei, Baohong Lv, Junjie Yan, Lingpeng Kong, and Yiran Zhong.
\newblock cos{F}ormer: Rethinking softmax in attention.
\newblock In {\em International Conference on Learning Representations}, 2022.

\bibitem[\protect\citeauthoryear{Qiu \bgroup \em et al.\egroup }{2020}]{qiu-etal-2020-blockwise}
Jiezhong Qiu, Hao Ma, Omer Levy, Wen-tau Yih, Sinong Wang, and Jie Tang.
\newblock Blockwise self-attention for long document understanding.
\newblock In {\em Findings of the Association for Computational Linguistics: EMNLP 2020}, pages 2555--2565, Online, November 2020. Association for Computational Linguistics.

\bibitem[\protect\citeauthoryear{Ratner \bgroup \em et al.\egroup }{2023}]{ratner-etal-2023-parallel}
Nir Ratner, Yoav Levine, Yonatan Belinkov, Ori Ram, Inbal Magar, Omri Abend, Ehud Karpas, Amnon Shashua, Kevin Leyton-Brown, and Yoav Shoham.
\newblock Parallel context windows for large language models.
\newblock In {\em Proceedings of the 61st Annual Meeting of the Association for Computational Linguistics (Volume 1: Long Papers)}, pages 6383--6402, Toronto, Canada, July 2023. Association for Computational Linguistics.

\bibitem[\protect\citeauthoryear{Roy \bgroup \em et al.\egroup }{2021}]{roy-etal-2021-efficient}
Aurko Roy, Mohammad Saffar, Ashish Vaswani, and David Grangier.
\newblock Efficient content-based sparse attention with routing transformers.
\newblock {\em Transactions of the Association for Computational Linguistics}, 9:53--68, 2021.

\bibitem[\protect\citeauthoryear{Saragadam \bgroup \em et al.\egroup }{2022}]{saragadam2022deeptensor}
Vishwanath Saragadam, Randall Balestriero, Ashok Veeraraghavan, and Richard~G Baraniuk.
\newblock Deep{T}ensor: Low-rank tensor decomposition with deep network priors.
\newblock {\em arXiv preprint arXiv:2204.03145}, 2022.

\bibitem[\protect\citeauthoryear{Shaw \bgroup \em et al.\egroup }{2018}]{shaw-etal-2018-self}
Peter Shaw, Jakob Uszkoreit, and Ashish Vaswani.
\newblock Self-attention with relative position representations.
\newblock In {\em Proceedings of the 2018 Conference of the North {A}merican Chapter of the Association for Computational Linguistics: Human Language Technologies, Volume 2 (Short Papers)}, pages 464--468, New Orleans, Louisiana, June 2018. Association for Computational Linguistics.

\bibitem[\protect\citeauthoryear{Shazeer}{2019}]{shazeer2019fast}
Noam Shazeer.
\newblock Fast transformer decoding: One write-head is all you need.
\newblock {\em arXiv preprint arXiv:1911.02150}, 2019.

\bibitem[\protect\citeauthoryear{Shen \bgroup \em et al.\egroup }{2023}]{shen2023study}
Kai Shen, Junliang Guo, Xu~Tan, Siliang Tang, Rui Wang, and Jiang Bian.
\newblock A study on {R}e{LU} and softmax in transformer.
\newblock {\em arXiv preprint arXiv:2302.06461}, 2023.

\bibitem[\protect\citeauthoryear{Su \bgroup \em et al.\egroup }{2024}]{su2024roformer}
Jianlin Su, Murtadha Ahmed, Yu~Lu, Shengfeng Pan, Wen Bo, and Yunfeng Liu.
\newblock Roformer: Enhanced transformer with rotary position embedding.
\newblock {\em Neurocomputing}, 568:127063, 2024.

\bibitem[\protect\citeauthoryear{Sun \bgroup \em et al.\egroup }{2023a}]{sun2023simple}
Mingjie Sun, Zhuang Liu, Anna Bair, and J~Zico Kolter.
\newblock A simple and effective pruning approach for large language models.
\newblock In {\em Workshop on Efficient Systems for Foundation Models @ ICML2023}, 2023.

\bibitem[\protect\citeauthoryear{Sun \bgroup \em et al.\egroup }{2023b}]{sun2023retentive}
Yutao Sun, Li~Dong, Shaohan Huang, Shuming Ma, Yuqing Xia, Jilong Xue, Jianyong Wang, and Furu Wei.
\newblock Retentive {N}etwork: A successor to transformer for large language models.
\newblock {\em arXiv preprint arXiv:2307.08621}, 2023.

\bibitem[\protect\citeauthoryear{Sun \bgroup \em et al.\egroup }{2023c}]{sun-etal-2023-length}
Yutao Sun, Li~Dong, Barun Patra, Shuming Ma, Shaohan Huang, Alon Benhaim, Vishrav Chaudhary, Xia Song, and Furu Wei.
\newblock A length-extrapolatable transformer.
\newblock In {\em Proceedings of the 61st Annual Meeting of the Association for Computational Linguistics (Volume 1: Long Papers)}, pages 14590--14604, Toronto, Canada, July 2023. Association for Computational Linguistics.

\bibitem[\protect\citeauthoryear{Touvron \bgroup \em et al.\egroup }{2023}]{touvron2023llama}
Hugo Touvron, Thibaut Lavril, Gautier Izacard, Xavier Martinet, Marie-Anne Lachaux, Timoth{\'e}e Lacroix, Baptiste Rozi{\`e}re, Naman Goyal, Eric Hambro, Faisal Azhar, Aurelien Rodriguez, Armand Joulin, Edouard Grave, and Guillaume Lample.
\newblock {LLaMA}: Open and efficient foundation language models.
\newblock {\em ArXiv}, abs/2302.13971, 2023.

\bibitem[\protect\citeauthoryear{Vaswani \bgroup \em et al.\egroup }{2017}]{NIPS2017_3f5ee243}
Ashish Vaswani, Noam Shazeer, Niki Parmar, Jakob Uszkoreit, Llion Jones, Aidan~N. Gomez, \L{}ukasz Kaiser, and Illia Polosukhin.
\newblock Attention is all you need.
\newblock In {\em Proceedings of the 31st International Conference on Neural Information Processing Systems}, NIPS'17, page 6000–6010, Red Hook, NY, USA, 2017.

\bibitem[\protect\citeauthoryear{Wan \bgroup \em et al.\egroup }{2023}]{wan2023efficient}
Zhongwei Wan, Xin Wang, Che Liu, Samiul Alam, Yu~Zheng, Zhongnan Qu, Shen Yan, Yi~Zhu, Quanlu Zhang, Mosharaf Chowdhury, et~al.
\newblock Efficient large language models: A survey.
\newblock {\em arXiv preprint arXiv:2312.03863}, 1, 2023.

\bibitem[\protect\citeauthoryear{Wang \bgroup \em et al.\egroup }{2020}]{wang2020linformer}
Sinong Wang, Belinda~Z Li, Madian Khabsa, Han Fang, and Hao Ma.
\newblock Linformer: Self-attention with linear complexity.
\newblock {\em arXiv preprint arXiv:2006.04768}, 2020.

\bibitem[\protect\citeauthoryear{Wang \bgroup \em et al.\egroup }{2023}]{wang2023investigating}
Xindi Wang, Yufei Wang, Can Xu, Xiubo Geng, Bowen Zhang, Chongyang Tao, Frank Rudzicz, Robert~E. Mercer, and Daxin Jiang.
\newblock Investigating the learning behaviour of in-context learning: A comparison with supervised learning.
\newblock {\em European Conference on Artificial Intelligence}, 2023.

\bibitem[\protect\citeauthoryear{Winata \bgroup \em et al.\egroup }{2020}]{winata2020lightweight}
Genta~Indra Winata, Samuel Cahyawijaya, Zhaojiang Lin, Zihan Liu, and Pascale Fung.
\newblock Lightweight and efficient end-to-end speech recognition using low-rank transformer.
\newblock In {\em ICASSP 2020 - 2020 IEEE International Conference on Acoustics, Speech and Signal Processing (ICASSP)}, pages 6144--6148, 2020.

\bibitem[\protect\citeauthoryear{Wu \bgroup \em et al.\egroup }{2021}]{NEURIPS2021_bcc0d400}
Haixu Wu, Jiehui Xu, Jianmin Wang, and Mingsheng Long.
\newblock Autoformer: Decomposition transformers with auto-correlation for long-term series forecasting.
\newblock In {\em Advances in Neural Information Processing Systems}, volume~34, pages 22419--22430, 2021.

\bibitem[\protect\citeauthoryear{Xia \bgroup \em et al.\egroup }{2024}]{xia2023sheared}
Mengzhou Xia, Tianyu Gao, Zhiyuan Zeng, and Danqi Chen.
\newblock Sheared {LL}a{MA}: Accelerating language model pre-training via structured pruning.
\newblock In {\em The Twelfth International Conference on Learning Representations}, 2024.

\bibitem[\protect\citeauthoryear{Xiao \bgroup \em et al.\egroup }{2024}]{xiao2023efficient}
Guangxuan Xiao, Yuandong Tian, Beidi Chen, Song Han, and Mike Lewis.
\newblock Efficient streaming language models with attention sinks.
\newblock In {\em The Twelfth International Conference on Learning Representations}, 2024.

\bibitem[\protect\citeauthoryear{Zhai \bgroup \em et al.\egroup }{2021}]{zhai2021attention}
Shuangfei Zhai, Walter Talbott, Nitish Srivastava, Chen Huang, Hanlin Goh, Ruixiang Zhang, and Josh Susskind.
\newblock An attention free transformer.
\newblock {\em arXiv preprint arXiv:2105.14103}, 2021.

\bibitem[\protect\citeauthoryear{Zhao \bgroup \em et al.\egroup }{2023}]{zhao2023survey}
Wayne~Xin Zhao, Kun Zhou, Junyi Li, Tianyi Tang, Xiaolei Wang, Yupeng Hou, Yingqian Min, Beichen Zhang, Junjie Zhang, Zican Dong, Yifan Du, Chen Yang, Yushuo Chen, Zhipeng Chen, Jinhao Jiang, Ruiyang Ren, Yifan Li, Xinyu Tang, Zikang Liu, Peiyu Liu, Jian-Yun Nie, and Ji-Rong Wen.
\newblock A survey of large language models, 2023.

\bibitem[\protect\citeauthoryear{Zhu \bgroup \em et al.\egroup }{2023}]{zhu2023survey}
Xunyu Zhu, Jian Li, Yong Liu, Can Ma, and Weiping Wang.
\newblock A survey on model compression for large language models.
\newblock {\em arXiv preprint arXiv:2308.07633}, 2023.

\bibitem[\protect\citeauthoryear{Zhuang \bgroup \em et al.\egroup }{2023}]{ijcai2023p0764}
Bohan Zhuang, Jing Liu, Zizheng Pan, Haoyu He, Yuetian Weng, and Chunhua Shen.
\newblock A survey on efficient training of transformers.
\newblock In {\em Proceedings of the Thirty-Second International Joint Conference on Artificial Intelligence, {IJCAI-23}}, pages 6823--6831. International Joint Conferences on Artificial Intelligence Organization, 8 2023.
\newblock Survey Track.

\end{thebibliography}

\end{document}